\documentclass[conference]{IEEEtran}
\IEEEoverridecommandlockouts
\usepackage{cite}
\usepackage{amsmath,amssymb,amsfonts}
\usepackage{algorithmic}
\usepackage{graphicx}
\usepackage{textcomp}
\usepackage{xcolor}
\def\BibTeX{{\rm B\kern-.05em{\sc i\kern-.025em b}\kern-.08em
    T\kern-.1667em\lower.7ex\hbox{E}\kern-.125emX}}
\begin{document}

\title{Biologically Realistic Dynamics for Nonlinear Classification in CMOS+X Neurons

}

\author{
\IEEEauthorblockN{
Steven Louis\IEEEauthorrefmark{1},
Hannah Bradley\IEEEauthorrefmark{2},
Artem Litvinenko\IEEEauthorrefmark{2},
Cody Trevillian\IEEEauthorrefmark{2},
Darrin Hanna\IEEEauthorrefmark{1},
Vasyl Tyberkevych\IEEEauthorrefmark{2}
}
\IEEEauthorblockA{\IEEEauthorrefmark{1}
\textit{Department of Electrical and Computer Engineering} 
}
\IEEEauthorblockA{\IEEEauthorrefmark{2}
\textit{Department of Physics} \\
\textit{Oakland University,} 
\textit{Rochester, MI  USA} \\
\textit{slouis@oakland.edu}
}
}

\maketitle

\begin{abstract}
Spiking neural networks encode information in spike timing and offer a pathway toward energy efficient artificial intelligence. 
However, a key challenge in spiking neural networks is realizing nonlinear and expressive computation in compact, energy-efficient hardware without relying on additional circuit complexity. 
In this work, we examine nonlinear computation in a CMOS+X spiking neuron implemented with a magnetic tunnel junction connected in series with an NMOS transistor. 
Circuit simulations of a multilayer network solving the XOR classification problem show that three intrinsic neuronal properties enable nonlinear behavior: threshold activation, response latency, and absolute refraction. 
Threshold activation determines which neurons participate in computation, response latency shifts spike timing, and absolute refraction suppresses subsequent spikes. 
These results show that magnetization dynamics of MTJ devices can support nonlinear computation in compact neuromorphic hardware.
\end{abstract}


\section{Introduction}

Artificial intelligence systems have advanced rapidly as neural network models have become larger and more complex \cite{bruscia2024overview}. 
These models are typically implemented on conventional CMOS hardware using arrays of artificial neurons that perform matrix operations and high-precision arithmetic \cite{zhu2023cmos}. 
Although effective, this approach requires substantial energy and is difficult to deploy on battery-powered edge devices where power consumption is constrained \cite{jouppi2017datacenter}.

Spiking neural networks (SNNs) offer a more energy-efficient computing paradigm. 
Unlike conventional neural networks that operate on continuous values, spiking neurons communicate using discrete events whose timing carries information \cite{maass1997networks, davies2018loihi}. 
Because computation occurs only when spikes are present, SNNs enable sparse event-driven processing well suited for low-power neuromorphic systems.

Recently, Louis et al. \cite{louis2025cmos+} demonstrated that a magnetic tunnel junction (MTJ) combined with a single NMOS transistor can function as a compact spiking neuron compatible with standard CMOS manufacturing processes. 
This NMOS+MTJ neuron, consisting of an NMOS transistor connected in series with an MTJ, is nearly identical to a conventional 1T-1MTJ MRAM cell and produces spiking behavior through intrinsic magnetization dynamics of the MTJ free layer rather than through complex control circuitry.

Beyond generating spikes, this device naturally reproduces several dynamical behaviors characteristic of biological neurons. 
These include all-or-nothing threshold activation, response latency, and refractory dynamics. 
In biological neural systems, such features play an essential role in shaping information processing \cite{squire2012fundamental, gerstner2002spiking}. 
However, many silicon neuron implementations reproduce only simplified versions of these behaviors, often omitting latency and refractory dynamics \cite{Frenkel2018}.

The NMOS+MTJ neuron is unusual in that these biologically realistic behaviors arise directly from device physics. 
In particular, magnetization dynamics of the MTJ free layer naturally produce threshold spiking, input-dependent response latency, and refractory behavior without requiring additional circuit elements \cite{louis2025cmos+}. 
These dynamics therefore provide computational mechanisms fundamentally different from those used in conventional CMOS neuron designs.

This NMOS+MTJ neuron also fits within a broader class of architectures often referred to as CMOS+X, in which conventional CMOS circuitry is combined with emerging device technologies that provide new device-level functionality. 
Examples of such devices include memristors, phase change materials, and spintronic elements such as MTJs. 
In this work, the MTJ provides the nonlinear dynamical behavior that enables spiking and temporal processing within an otherwise standard CMOS circuit.

While the NMOS+MTJ neuron and its learning behavior were previously demonstrated \cite{louis2025cmos+}, the computational role of these intrinsic neuronal dynamics has not been examined in detail. 
In particular, it remains unclear how threshold activation, response latency, and refractory dynamics contribute to nonlinear computation in such networks.

In this work, we examine how these three biologically realistic properties enable nonlinear classification in NMOS+MTJ spiking neural networks. 
Using the XOR problem as a benchmark, we analyze how threshold activation determines which neurons participate in computation, how response latency encodes information in spike timing, and how absolute refraction prevents subsequent spikes from altering the output spike already in progress.

By connecting device-level magnetization dynamics to network-level behavior, this work demonstrates that biologically realistic neuronal features can serve as intrinsic computational resources in CMOS+X neuromorphic systems. 
In this architecture, nonlinear classification emerges directly from device physics rather than from additional circuit complexity.

%new version
\section{The NMOS+MTJ Spiking Neuron Structure}

\begin{figure}
\centering
\includegraphics{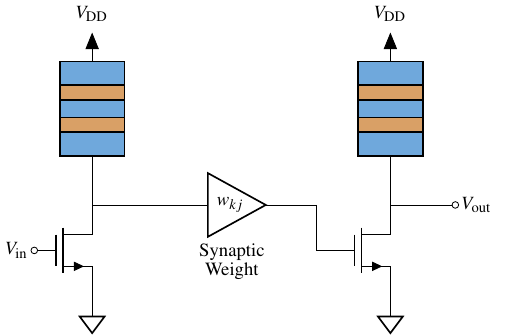}
\caption{Electrical schematic illustrating two NMOS+MTJ neurons connected by a synaptic weight. 
Each neuron consists of a magnetic tunnel junction (MTJ), represented by the blue and orange layers, connected in series with an NMOS transistor. 
The synaptic connection is implemented with a voltage amplifier whose gain represents the synaptic weight $w_{kj}$ that controls how activity from one neuron influences the next neuron. 
$V_\textrm{in}$ is the input voltage applied to the first neuron, $V_\textrm{out}$ is the output voltage of the second neuron, and $V_\textrm{DD}$ is the supply voltage.}\vspace{-4mm}
\label{fig_neuron}
\end{figure}

The artificial neuron in this work is the NMOS+MTJ circuit introduced in \cite{louis2025cmos+} and derived from the MTJ neuron concept and circuit model described in \cite{mtjneuron, louis2025physics}. 
As illustrated in Fig.~\ref{fig_neuron}, the neuron consists of a magnetic tunnel junction (MTJ) connected in series with an NMOS transistor. 
This configuration resembles a one-transistor-one-MTJ (1T-1MTJ) MRAM cell and is compatible with existing CMOS manufacturing processes and embedded MRAM technologies~\cite{everspinEverspinMRAM}.

Spiking behavior in the NMOS+MTJ neuron arises from magnetization dynamics in the MTJ free layer rather than from additional control circuitry. 
Prior work showed that the MTJ resistance depends on free layer orientation and that current through the device generates spin-transfer torque capable of producing a spiking response \cite{mtjneuron, louis2025cmos+}. 
When combined in series with an NMOS transistor, this resistance change appears as a voltage spike at the output \cite{louis2025cmos+}. 
The present work demonstrates that this neuron naturally exhibits threshold activation, response latency, and refractory behavior, as examined below.

NMOS+MTJ neurons can be connected through analog synapses implemented as tunable voltage amplifiers. 
As shown in Fig.~\ref{fig_neuron}, the output of one neuron drives the input of another through an amplifier representing the synaptic weight. 
The amplifier gain sets the connection strength, allowing spikes to propagate between neurons and trigger downstream spiking with latency determined by the synaptic gain \cite{louis2025cmos+}.

Although the NMOS transistor controls the current amplitude, the nonlinear spiking behavior originates from the MTJ, whose magnetization dynamics determine whether and when a resistance transition occurs. 
In the absence of MTJ dynamics, the circuit reduces to a linear resistive element and does not exhibit the biological features covered in this work.

Importantly, threshold activation, response latency, synaptic integration, and both absolute and relative refraction arise naturally from magnetization dynamics of the MTJ free layer without additional circuitry \cite{louis2025cmos+}. 
These dynamics form the basis for the nonlinear network behavior examined in this work.

\section{XOR Classification Network}

The XOR function is a standard nonlinear benchmark that cannot be solved by a single-layer perceptron and therefore requires multilayer processing with nonlinear activation, making it an ideal, simple test case for our work.  
To enable gradient-based training, the network is constrained to a single-spike regime in which each neuron emits at most one spike per input, allowing spike timing to be used for backpropagation.

\begin{figure}
\centering
\includegraphics{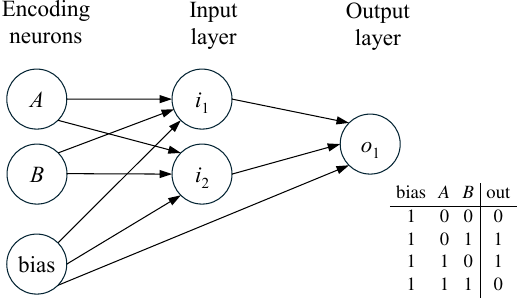}
\caption{Architecture of the NMOS+MTJ neural network used for XOR classification. 
Encoding neurons $A$ and $B$ drive input-layer neurons $i_1$ and $i_2$, which connect to output neuron $o_1$ through synaptic weights. 
A bias neuron provides a constant input to the network. 
The XOR truth table showing the bias input and target output.}\vspace{-4mm}
\label{fig_xor}
\end{figure}

Figure~\ref{fig_xor} shows the NMOS+MTJ neural network used for XOR classification together with the corresponding truth table.
The network consists of encoding neurons $A$ and $B$ that generate input spikes corresponding to the truth table values, input-layer neurons $i_1$ and $i_2$, and output neuron $o_1$.
A bias neuron provides a constant input spikes to the network.

Synaptic connections are implemented using the tunable voltage amplifier model described in Section II. 
The amplifier gain corresponds to a synaptic weight $w_{kj}$ connecting neuron $j$ to neuron $k$. 
These weights determine how strongly spikes from one neuron influence the next.

During operation, spikes generated by neurons $i_1$ and $i_2$ propagate through the synaptic connections to the output neuron $o_1$. 
The spike timing of $o_1$ encodes the XOR output. 
Following the convention introduced in \cite{louis2025cmos+}, a spike of $o_1$ at $t = 2\,\text{ns}$ represents 0, while a spike at $t = 2.5\,\text{ns}$ represents 1.

The NMOS+MTJ network was trained using a spike-timing-based gradient descent algorithm implemented entirely in analog circuitry, with the learning rule derived from the loss function
\begin{equation}
L = \frac{1}{2}(t_{\text{actual}} - t_{\text{desired}})^2
\end{equation}
where $t_{\text{actual}}$ is the output spike time of neuron $o_1$ and $t_{\text{desired}}$ is the target spike time \cite{louis2025cmos+}. 
Weight updates follow
\begin{equation}
\Delta w_{kj} = -\eta \frac{dL}{dt_k}\frac{\partial t_k}{\partial w_{k,j}},
\end{equation}
where $\eta$ is the learning rate and $t_k$ is the firing time of the presynaptic neuron.
During training, synaptic weights are adjusted to minimize the difference between the observed and desired output spike times specified by the XOR truth table.

The reader is directed to Section V and Appendix B of \cite{louis2025cmos+} for a complete derivation of the learning rule and its analog circuit implementation. 
Here, we focus instead on how intrinsic biological features of the NMOS+MTJ neuron enable nonlinear XOR classification.

\section{Biological Features Enabling XOR Classification}

In the NMOS+MTJ neural network, nonlinear capability arises from three biologically realistic features of the neuron: threshold activation, response latency, and absolute refraction. 
Together, these properties create timing-dependent dynamics that allow the network to distinguish between input patterns that cannot be separated by linear systems.

\subsection{Threshold Activation}

The first mechanism enabling nonlinear behavior in the network is threshold activation. 
Biological neurons exhibit an all-or-nothing response in which spiking occurs only when the input exceeds a critical threshold. 
A similar behavior emerges naturally in the NMOS+MTJ neuron through the magnetization dynamics of the MTJ free layer.

\begin{figure}[t]
\centering
\includegraphics{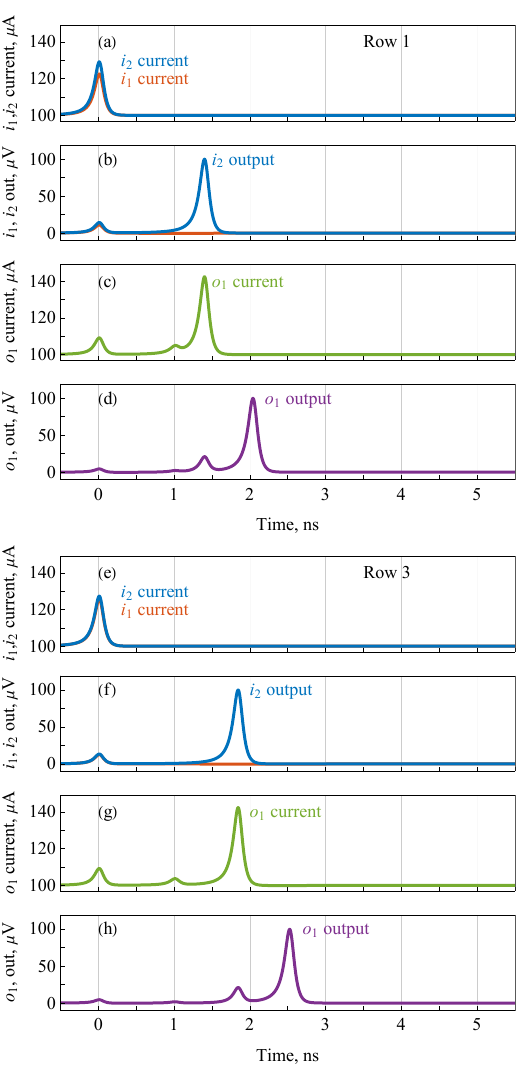}
\caption{Simulation results of rows 1 and 3 of the XOR truth table.
(a,e) Input currents supplied to neurons $i_1$ and $i_2$.
(b, f) Voltage spikes of neurons $i_1$ and $i_2$.
(c, g) Current supplied to output neuron $o_1$.
(d, h) Voltage spike of output neuron $o_1$.
}\vspace{-4mm}
\label{fig_latency}
\end{figure}

This effect is illustrated in Fig.~\ref{fig_latency}(a-d), which show the network response for row 1 of the XOR truth table, corresponding to the input combination $(A,B)=(0,0)$. 
Due to differences in the synaptic weights connecting the bias neuron to the input-layer neurons, the currents delivered to neurons $i_1$ and $i_2$ are slightly different. 
Under these conditions, neuron $i_2$ receives a current that exceeds the firing threshold, while neuron $i_1$ remains below threshold and does not spike. 
As a result, only neuron $i_2$ generates an output spike, which is transmitted to the output neuron $o_1$.

The output spike of neuron $o_1$ occurs at approximately $t=2$~ns, as shown in Fig.~\ref{fig_latency}(d). 
In this network, the spike timing encodes the logical output of the XOR operation: a spike at $t=2$~ns represents a logical 0. 
Thus, threshold activation ensures that only inputs that exceed the firing condition propagate through the network, establishing the first stage of nonlinear processing.

The timing of the output spike is set by the time required for the input current to drive the MTJ free-layer magnetization across the switching threshold. 
Larger input currents produce stronger spin-transfer torque and faster switching, while currents closer to threshold result in slower magnetization dynamics and delayed spikes.

\subsection{Response Latency}

The second mechanism enabling nonlinear computation is response latency. 
In biological neurons, there is a finite delay between the arrival of an input and the generation of an output spike. 
This delay depends on the input strength, with stronger inputs producing shorter latencies. 
The NMOS+MTJ neuron exhibits this behavior due to the dependence of magnetization dynamics on the magnitude of the spin-transfer torque.

The effect of response latency is demonstrated in Fig.~\ref{fig_latency}(e-h), which show the network response for row 3 of the XOR truth table corresponding to the input $(A,B)=(1,0)$. 
In this case, the input-layer neurons receive different currents due to the trained synaptic weights connecting the encoding neurons to the network. 
Under these conditions, neuron $i_2$ receives a current that exceeds the firing threshold, while neuron $i_1$ receives a slightly smaller current that remains below threshold and therefore does not spike.

Compared with the response shown in Fig.~\ref{fig_latency}(b), the spike generated by neuron $i_2$ in Fig.~\ref{fig_latency}(f) occurs with a longer delay between the input current and the output voltage spike. 
This increased latency occurs because the input current is closer to the threshold condition, which slows the magnetization dynamics responsible for spike generation.

The increased latency shifts the timing of the spike received by the output neuron $o_1$, causing it to fire at approximately $t=2.5$~ns as shown in Fig.~\ref{fig_latency}(h). 
In the network encoding scheme, this delayed spike represents a logical 1. 
Thus, response latency provides a timing-based nonlinear mechanism that allows the network to distinguish between different input combinations.

\subsection{Absolute Refraction}

The third biologically realistic feature enabling XOR classification is absolute refraction. 
After generating a spike, biological neurons enter a refractory period during which they cannot fire again, regardless of input strength. 
This behavior also emerges in the NMOS+MTJ neuron as a consequence of the transient dynamics of the MTJ free-layer magnetization.

\begin{figure}
\centering
\includegraphics{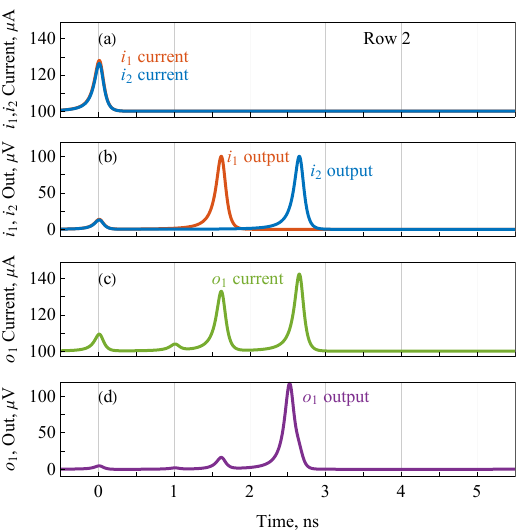}
\caption{Simulation result of row 2 of the XOR truth table.
(a) Input currents supplied to neurons $i_1$ and $i_2$.
(b) Voltage spikes of neurons $i_1$ and $i_2$.
(c) Current supplied to output neuron $o_1$.
(d) Voltage spike of output neuron $o_1$.}\vspace{-4mm}
\label{fig_refraction}
\end{figure}

The role of absolute refraction in the XOR network is illustrated in Fig.~\ref{fig_refraction}. 
In this case, both input-layer neurons $i_1$ and $i_2$ generate spikes that are transmitted to the output neuron $o_1$. 
The first incoming spike causes $o_1$ to fire at approximately $t=2.5$~ns, producing the correct logical output.

% However, when the second spike arrives shortly afterward, neuron $o_1$ is still within its refractory period and therefore cannot generate a second spike. 

As shown in Fig.~\ref{fig_refraction}(c), a second input current pulse exceeding the firing threshold is applied to neuron $o_1$. 
However, this pulse overlaps in time with the output spike, neuron $o_1$ which is still within its refractory period and the ongoing magnetization dynamics are not disturbed, and a second spike is not generated.
This suppression of additional spikes prevents the neuron from producing multiple outputs in response to closely spaced inputs.

The refractory behavior effectively implements a time-domain winner-take-all mechanism in which the earliest spike determines the network response. 
Without this mechanism, the output neuron could produce multiple spikes and fail to generate the correct XOR classification.

The fourth XOR, case $(A,B)=(1,1)$, which is not shown, produces a response similar to $(0,0)$, in which the output neuron fires at $t=2$~ns, with the timing determined by response latency.

Together, threshold activation, response latency, and absolute refraction form intrinsic dynamical mechanisms that enable nonlinear computation in the NMOS+MTJ SNN.

\section{Conclusion}

In this work, we examined how the NMOS+MTJ neuron, previously presented in \cite{louis2025cmos+}, leverages three biologically realistic features to perform nonlinear classification.
Through simulation results for the XOR truth table, we demonstrated that threshold activation, response latency, and absolute refraction each contribute to the ability of the network to solve this nonlinear classification problem.

Threshold activation ensures that only inputs exceeding a critical level produce spikes. 
Response latency introduces input-dependent timing variations that encode information in spike timing. 
Absolute refraction prevents additional spikes from altering the output spike already in progress. 
Together these mechanisms create timing-dependent dynamics that enable nonlinear computation.

Our analysis confirms that the NMOS+MTJ neural network successfully classifies the XOR function by harnessing the combined temporal dynamics of threshold activation, response latency, and absolute refraction. 
These results demonstrate that biologically realistic neural behaviors, rather than serving merely as biological analogies, serve as fundamental mechanisms enabling nonlinear computation in spiking neural networks.

% \section*{Acknowledgment}

% The preferred spelling of the word ``acknowledgment'' in America is without 
% an ``e'' after the ``g''. Avoid the stilted expression ``one of us (R. B. 
% G.) thanks $\ldots$''. Instead, try ``R. B. G. thanks$\ldots$''. Put sponsor 
% acknowledgments in the unnumbered footnote on the first page.

\bibliography{references}

@article{louis2025cmos+,
  title={A CMOS+ X Spiking Neuron With On-Chip Machine Learning},
  author={Louis, Steven and Abramson, Matthew Blake and Bradley, Hannah and Trevillian, Cody and Nelson, Gene David and Slavin, Andrei and Litvinenko, Artem and Gorski, Jason and Krivorotov, Ilya N and Hanna, Darrin and others},
  journal={arXiv preprint arXiv:2512.03966},
  year={2025}
}

@ARTICLE{Frenkel2018,
  author={Frenkel, Charlotte and Lefebvre, Martin and Legat, Jean-Didier and Bol, David},
  journal={IEEE Transactions on Biomedical Circuits and Systems}, 
  title={A 0.086-mm$^2$ 12.7-pJ/SOP 64k-Synapse 256-Neuron Online-Learning Digital Spiking Neuromorphic Processor in 28-nm CMOS}, 
  year={2019},
  volume={13},
  number={1},
  pages={145-158},
  doi={10.1109/TBCAS.2018.2880425}}

@book{squire2012fundamental,
  title={Fundamental neuroscience},
  author={Squire, Larry and Berg, Darwin and Bloom, Floyd E and Du Lac, Sascha and Ghosh, Anirvan and Spitzer, Nicholas C},
  year={2012},
  publisher={Academic press}
}

@inproceedings{jouppi2017datacenter,
  title={In-datacenter performance analysis of a tensor processing unit},
  author={Jouppi, Norman P and Young, Cliff and Patil, Nishant and Patterson, David and Agrawal, Gaurav and Bajwa, Raminder and Bates, Sarah and Bhatia, Suresh and Boden, Nan and Borchers, Al and others},
  booktitle={Proceedings of the 44th annual international symposium on computer architecture},
  pages={1--12},
  year={2017}
}

@article{maass1997networks,
  title={Networks of spiking neurons: the third generation of neural network models},
  author={Maass, Wolfgang},
  journal={Neural networks},
  volume={10},
  number={9},
  pages={1659--1671},
  year={1997},
  publisher={Elsevier}
}

@article{davies2018loihi,
  title={Loihi: A neuromorphic manycore processor with on-chip learning},
  author={Davies, Mike and Srinivasa, Narayan and Lin, Tsung-Han and Chinya, Gautham and Cao, Yongqiang and Choday, Sri Harsha and Dimou, Georgios and Joshi, Prasad and Imam, Nabil and Jain, Shweta and others},
  journal={Ieee micro},
  volume={38},
  number={1},
  pages={82--99},
  year={2018},
  publisher={IEEE}
}

@book{gerstner2002spiking,
  title={Spiking neuron models: Single neurons, populations, plasticity},
  author={Gerstner, Wulfram and Kistler, Werner M},
  year={2002},
  publisher={Cambridge university press}
}

@article{louis2025physics,
  title={A Physics-Based Circuit Model for Magnetic Tunnel Junctions},
  author={Louis, Steven and Bradley, Hannah and Litvinenko, Artem and Tyberkevych, Vasyl},
  journal={IEEE Magnetics Letters},
  year={2025},
  publisher={IEEE}
}

@misc{everspinEverspinMRAM,
	author = {},
	title = {{E}verspin | {T}he {M}{R}{A}{M} {C}ompany --- everspin.com},
	howpublished = {\url{https://www.everspin.com/}},
	year = {},
	note = {[Accessed 01-03-2026]},
}

@article{zhu2023cmos,
  title={CMOS-compatible neuromorphic devices for neuromorphic perception and computing: a review},
  author={Zhu, Yixin and Mao, Huiwu and Zhu, Ying and Wang, Xiangjing and Fu, Chuanyu and Ke, Shuo and Wan, Changjin and Wan, Qing},
  journal={International Journal of Extreme Manufacturing},
  volume={5},
  number={4},
  pages={042010},
  year={2023},
  publisher={IOP Publishing}
}

@inproceedings{bruscia2024overview,
  title={An overview on large language models across key domains: a systematic review},
  author={Bruscia, Mattia and Manduzio, Graziano A and Galatolo, Federico A and Cimino, Mario GCA and Greco, Alberto and Cominelli, Lorenzo and Scilingo, Enzo Pasquale},
  booktitle={2024 IEEE International Conference on Metrology for eXtended Reality, Artificial Intelligence and Neural Engineering (MetroXRAINE)},
  pages={125--130},
  year={2024},
  organization={IEEE}
}

@ARTICLE{mtjneuron,
  author={Louis, Steven and Bradley, Hannah and Trevillian, Cody and Slavin, Andrei and Tyberkevych, Vasyl},
  journal={IEEE Magnetics Letters}, 
  title={Spintronic Neuron Using a Magnetic Tunnel Junction for Low-Power Neuromorphic Computing}, 
  year={2024},
  volume={15},
  number={},
  pages={1-5},
  doi={10.1109/LMAG.2024.3484957}}
\bibliographystyle{IEEEtran}

\end{document}